\documentclass[runningheads]{llncs}

\usepackage{eccv}

\usepackage{eccvabbrv}

\usepackage{graphicx}
\usepackage{booktabs}

\usepackage[accsupp]{axessibility}  
\usepackage{enumitem}
\newlist{steps}{enumerate}{1}
\usepackage{amssymb}
\usepackage{pifont}
\newcommand{\xmark}{\ding{55}}
\newcommand{\cmark}{\ding{51}}
\usepackage{wrapfig}

\usepackage{hyperref}

\usepackage{orcidlink}

\begin{document}

\title{Constructing Concept-based Models to Mitigate Spurious Correlations with Minimal Human Effort} 

\titlerunning{Constructing Concept-based Models with Minimal Human Effort}

\author{Jeeyung Kim\and Ze Wang \and
Qiang Qiu}

\authorrunning{J.~Kim et al.}

\institute{Purdue University, West Lafayette, USA\\\email{\{jkim17, wang5026, qqiu\}@purdue.edu} 
}

\maketitle

\begin{abstract}
Enhancing model interpretability can address spurious correlations by revealing how models draw their predictions. 
Concept Bottleneck Models (CBMs) can provide a principled way of disclosing and guiding model behaviors through human-understandable concepts, albeit at a high cost of human efforts in data annotation.
In this paper, we leverage a synergy of multiple foundation models to construct CBMs with nearly no human effort.
We discover undesirable biases in CBMs built on pre-trained models and propose a novel framework designed to exploit pre-trained models while being immune to these biases, thereby reducing vulnerability to spurious correlations.
Specifically, our method offers a seamless pipeline that adopts foundation models for assessing potential spurious correlations in datasets, annotating concepts for images, and refining the annotations for improved robustness. 
We evaluate the proposed method on multiple datasets, and the results demonstrate its effectiveness in reducing model reliance on spurious correlations while preserving its interpretability.
  \keywords{Multimodal Large Language Model for Data Annotation \and Spurious Correlations \and Concept Bottleneck Model}
\end{abstract}

\section{Introduction}
\label{sec:intro}

Deep learning models excel in various vision tasks, such as image classification, by learning from massive training data. 
However, the \textit{black-box} nature of deep learning models makes them vulnerable to spurious correlations residing in training data.  
Specifically, models may rely on spurious relationships for predictions, often unaware of having learned such flawed correlations. Such learned spurious correlations can hardly be diagnosed and eliminated after model training. 

A promising direction to address this issue is to improve model interpretability by using human-comprehensible \textit{concepts} introduced in prior studies~\cite{kim2018interpretability, koh2020concept}.
These methods elucidate how a set of pre-defined \textit{concepts} contributes to model prediction. 
In particular, \cite{koh2020concept} introduces Concept Bottleneck Models (CBMs), which encode inputs based on the presenting concepts and use these concepts to draw predictions.
Moreover, \cite{abid2022meaningfully, wu2023discover} demonstrate using human-understandable \textit{concepts} in identifying and addressing spurious correlations.
Despite their advantages, such methods require concept annotations of the images, which demand considerably greater human effort compared to black-box models.

Recent advancements in large foundation models, such as CLIP~\cite{radford2021learning} and GPT-3~\cite{brown2020language}, have substantially reduced the cost required to construct CBMs.
\cite{yuksekgonul2022post, oikarinen2023label} present methods to represent concepts using features from CLIP to ease CBM construction.
\cite{oikarinen2023label} uses GPT-3 to collect relevant concepts for specific classification tasks.
Moreover, \cite{yuksekgonul2022post} proposes a technique to mitigate spurious correlations in CBMs by pruning classifier weights tied with misleading concepts, exhibiting the capability of CBMs to reduce spurious correlations. 
However, selecting weights to prune still requires human expertise. In addition, as demonstrated in \cref{sec:post_hoc_issue}, such CBMs are limited in addressing spurious correlations as they tend to overlook the biases inherent in large foundation models like CLIP.

To overcome the aforementioned shortcomings, we present a framework for constructing CBMs to effectively tackle spurious correlations at a low cost, leveraging the capacities of multiple foundation models. 
Specifically, the proposed framework accomplishes CBM constructions in three stages. 
In the first stage, we employ a multimodal large language model (MLLM) and a large language model (LLM) to summarize datasets and create comprehensive concept pools. The LLM identifies all the potentially helpful visual concepts for each dataset. Considering that some concepts may be associated with spurious correlations, automatic concept filtering with MLLM is applied to identify and remove those tied to potential spurious correlations within each dataset.
In the second stage, automatic concept annotations are performed using MLLM based on the filtered concept list. Inspired by the observed vulnerability of the CBMs built on pre-trained models \cite{yuksekgonul2022post, oikarinen2023label} to spurious correlations, as showcased in \cref{sec:post_hoc_issue}, we obtain binary annotations through foundation models rather than using potentially biased raw representations from the pre-trained models. The resulting CBMs effectively minimize inheriting unintended biases from pre-trained models.
Despite the remarkable capabilities of MLLMs on various tasks, we recognize the inferior concept annotation quality of MLLMs compared to human experts. To bridge the gap in concept annotation accuracy, we introduce an optional third step for annotation refinement using chains of vision foundation models, improving the reliability of the concept annotation with MLLMs with minimal human effort.
Applying this framework, we develop CBMs and showcase their efficacy in addressing spurious correlations with little to no reliance on human labor in real-world challenges.
\begin{figure*}
\centering
\includegraphics[width=0.6\linewidth]{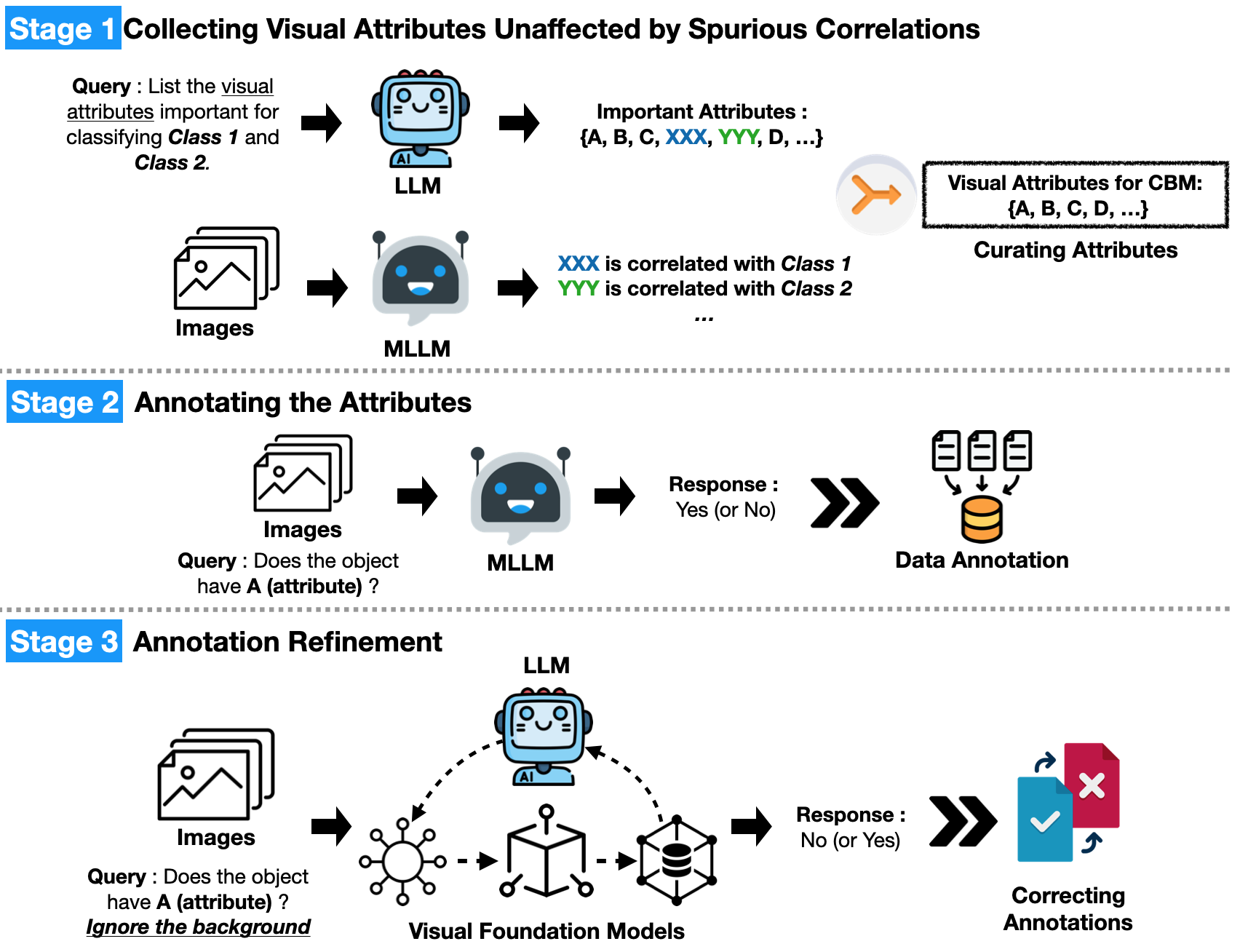}
    \caption{The framework comprises three stages, where we leverage foundation models to minimize human effort in building CBMs.  First, we collect a concept pool of visual attributes necessary for classification with automatic concept filtering using an MLLM. Second, we annotate the concepts by querying a MLLM. Third, we optionally refine the annotations using vision foundation models to improve the accuracy of the concept annotations by MLLMs. Ultimately, based on the obtained concept annotations, we construct a LLaVA-based CBM.}
    \label{fig:framework}
\end{figure*}

\section{Related Works}
\textbf{Concept-based Models.}
Using human-understandable \textit{concepts} to guide and interpret model behavior has been investigated in~\cite{koh2020concept, kim2018interpretability, yeh2020completeness, abid2022meaningfully, wu2023discover, chen2020concept}.
Particularly, \cite{koh2020concept} introduce CBMs mapping individual neurons in an intermediate layer to concepts, enabling human intervention at test time to improve accuracy. \cite{kim2018interpretability} proposes Concept Activation Vectors (CAVs), aligned with feature space of an intermediate layer to assess the influence of concepts on model predictions.
However, the enhanced interpretability of these models comes at the cost and reduced performance compared to black-box models.
While recent studies~\cite{yuksekgonul2022post, oikarinen2023label} leverage the capabilities of pre-trained models such as CLIP and GPT-3 to reduce the cost for building CBMs and matching black-box performance, its potential to mitigate spurious correlations in real-world scenarios remains unexplored.

\noindent\textbf{Improving robustness to spurious correlations.}
Previous studies~\cite{zhang2022correct, liu2021just, kirichenko2022last, sagawa2019distributionally, ye2023freeze, yang2023mitigating, wu2023discover,yang2023mitigating} improve the robustness of black-box models to spurious correlations, but their lack of interpretability hinders deep understanding of model behaviors.
Moreover, the approaches proposed in~\cite{sagawa2019distributionally, zhang2020coping, sagawa2020investigation} require group annotations that are hard to obtain.
To address spurious correlations through model behavior interpretation, prior studies~\cite{abid2022meaningfully, wu2023discover, bontempelli2022concept, yuksekgonul2022post} employ {\it concepts} and reveal spurious correlations in a human-interpretable manner.
Particularly, \cite{abid2022meaningfully, wu2023discover} utilizes CAVs to elucidate spurious correlations within models and \cite{wu2023discover} further mitigates the discovered biases through concept-aware interventions. 
However, generating CAVs is expensive as it requires concept annotations of images.
In addition, \cite{yuksekgonul2022post} introduces CBMs built on the pre-trained model and proposes a weight-pruning method for mitigating spurious correlations. Nevertheless, it is not applicable to large-scale setups and does not adequately address model biases inheriting from the pre-trained models.

\noindent\textbf{Foundation Models.}
Due to the impressive capabilities of LLMs, numerous studies have focused on extending LLMs' capabilities to incorporate multimodal inputs such as images and audios~\cite{alayrac2022flamingo, awadalla2023openflamingo, li2023blip, li2023otter, peng2023kosmos, zhu2023minigpt, wu2023next, zeng2022socratic}. 
LLaVA~\cite{liu2023visual, liu2023improved} stands out for its exceptional performance in visual and language understanding among open-source MLLMs.
In addition, \cite{yao2022react} demonstrates how to use LLMs for interactive decision-making by leveraging external tools (\eg search engine, foundation models). 
They propose effective strategies for planning and activating these tools using LLMs. 
Similarly, recent studies~\cite{chen2022visualgpt, chen2023llava, yang2023mm,gupta2023visual} introduce approaches to solve diverse vision tasks, such as visual math reasoning, image spatial understanding and image editing, by exploiting different vision foundation models (VFMs).
A chain of VFMs, activated by LLMs (or MLLMs) for specific sub-tasks, enhances the ability to accomplish the intended vision task. 
Furthermore, while recent advancements in LLMs prompted many studies to use them for generating training data~\cite{wang2023noise, rafailov2024direct, yan2023learning,yang2023language, li2024desco}, the use of MLLMs for data annotation remains largely unexplored.
To our knowledge, we are the first to investigate the use of MLLMs for data annotation.

\section{Preliminaries}
\label{sec:problem_setup}
CBMs are constructed with human-comprehensible \textit{concepts}.
For example, in the CUB dataset~\cite{wah2011caltech}, which includes 200 bird species, \textit{black wing color} and \textit{cone bill shape} are used as concepts to build CBMs. Note that we interchangeably use \textit{attribute} and \textit{concept} in the following sections.
CBMs comprise two-stages: a concept model ($g$) and a classifier ($f$).
The concept model predicts concept presence in the input, while the classifier predicts labels based on \textit{concept representation} ($\hat g (\mathbf{x})$), output of the concept model, where $g : \mathbb{R}^d \rightarrow \mathbb{R}^m$, $f : \mathbb{R}^m \rightarrow \mathbb{R}^l$, $d$ denotes the dimension of inputs, $m$ denotes the number of concepts used and $l$ indicates the number of class.
In this work, we only focus on cases where $f$ and $g$ are independently trained~\cite{koh2020concept}.

In \cite{koh2020concept}, a concept model is trained to predict a binary annotation for each concept. This setup requires datasets comprising tuples of images, labels, and binary concept annotations $\{(\mathbf{x}^{(j)}, \mathbf{y}^{(j)}, \mathbf{c}^{(j)})\}_{j=1}^n$, where $\mathbf{x} \in \mathbb{R}^d$, $\mathbf{y} \in \mathbb{R}^l$ and $\mathbf{c} \in \{0,1\}^m$. 
We dub this approach as \textit{Annotation-based CBM}.
Recently, Post-hoc CBM~\cite{yuksekgonul2022post} and Label-free CBM~\cite{oikarinen2023label} suggest building CBMs using CLIP text encoder, removing the need to train a concept model.
Particularly, \cite{yuksekgonul2022post} obtains $\hat g_i (\mathbf{x})$ by projecting image embedding ($h(\mathbf{x})$) to the corresponding text embedding from CLIP ($\mathcal{C}_i$), such that  $\hat g_{i} (\mathbf{x}) = \frac{ \langle h(\mathbf{x}), \mathcal{C}_i\rangle}{\|\mathcal{C}_i\|_2^2} \in \mathbb{R} $, where $i=1,2, \cdots, m$, $h(\mathbf{x}) \in \mathbb{R}^e$, $\mathcal{C} \in \mathbb{R}^{m \times e}$ denotes concept vectors from CLIP text encoder and $e$ is the size of the embedding space.
We term these approaches as \textit{CLIP-based CBM}.

We categorize CBMs into soft CBMs and hard CBMs based on how concepts $\mathbf{c}$ are modeled, following \cite{havasi2022addressing}.
A soft CBM characterizes $\mathbf{c}_i$ as a real number representing the degree of relatedness or presence of a concept.
In contrast, a hard CBM represents $\mathbf{c}_i$ with a binary value, indicating the presence or absence of each concept in input.
Annotation-based CBMs align with hard CBMs, while CLIP-based CBMs fall into the category of soft CBMs.

\section{Limitation of using CLIP-based CBMs to Alleviate Spurious Correlation}
\label{sec:post_hoc_issue}

Compared to end-to-end models, CBMs have the potential to reduce spurious correlations by filtering out concepts identified to contribute to spurious correlations with the aid of human expertise. In this section, we highlight key observations suggesting that despite careful concept selection, \textit{CLIP-based CBMs} can still be compromised by spurious correlations due to inherent biases in CLIP.

We experiment on Waterbirds~\cite{sagawa2019distributionally}, a benchmark synthesized by pasting birds from CUB~\cite{wah2011caltech} onto background images from Places~\cite{zhou2017places}. 
The dataset exhibits spurious correlations, with \textit{landbirds} predominantly appearing against \textit{land} backgrounds and \textit{waterbirds} mostly set against \textit{water} backgrounds.
We use the 112 visual concepts provided with CUB in \cite{koh2020concept} for constructing CBMs, so that they only consist of attributes related to the appearances of birds, not background. 
To assess the robustness of \textit{concept representation} ($\hat g(\mathbf{x})$) from concept models of CBMs against spurious correlations, we train an additional classifier, $f'(\cdot)$, which takes $\hat g(\mathbf{x})$ as input and predicts the labels associated with spurious correlations, \textit{land} and \textit{water}.
Table \ref{tab:spurious_test} shows that the classifier ($f'(\cdot)$) with concept representations of the CLIP-based CBM (post-hoc CBM) attaining an accuracy largely exceeding random guesses (50\%). 
This suggests that even only with concepts irrelevant to spurious correlations, $\hat g(\mathbf{x})$ of the CLIP-based CBM can still be noticeably biased towards the spurious elements. 
Conversely, the accuracy of a classifier built on $\hat g(\mathbf{x})$ of the Annotation-based CBM~\cite{koh2020concept} is as low as random guesses, implying that $\hat g(\mathbf{x})$ from the Annotation-based CBM are nearly free from biases leading to spurious correlations.
\begin{table}
  \centering
  \caption{Background (\textit{land}, \textit{water}) prediction accuracy on Waterbirds. The higher accuracy achieved by using concept representations from the CLIP-based CBM (Post-hoc CBM) suggests that the representations contain biases towards spurious correlations. }
  \label{tab:spurious_test}
  \begin{tabular}{@{}lcc@{}}
    \toprule
    Model & Background Acc.   \\
    \midrule
    Standard ResNet-50\cite{he2016deep} & 90.1\%\\
    Post-hoc CBM~\cite{yuksekgonul2022post} &  72.8\% \\
    Annotation-based CBM\cite{koh2020concept} &  53.0\%  \\
    \bottomrule
  \end{tabular}
\end{table}

There are two possible reasons why concept representations from CLIP-based CBMs exhibit biases, in contrast to those from Annotation-based CBMs. 
The first reason can relate to the inherent biases residing in CLIP.
CLIP-based CBMs generate concept vectors ($\mathcal{C}$) using CLIP encoders, which can be affected by biases in pre-training data.
For example, during the training phase of CLIP, visual attributes of \textit{waterbirds} can often appear against oceanic backgrounds, whereas \textit{landbirds}'s attributes more commonly appear with forest environments. 
This can accidentally encode a bias within the encoder that CLIP-based CBMs later use.
Second, the difference can stem from the types of CBMs. 
As mentioned in \cref{sec:problem_setup}, CLIP-based CBMs belongs to soft CBMs, while Annotation-based CBMs are hard CBMs.
\cite{havasi2022addressing} shows the potential risk that soft CBMs can allow the leakage of unintended information from the concept model to the classifier in contrast to hard CBMs. 
This leakage in soft CBMs may produce biased features, rendering it vulnerable to spurious correlations. 

\section{Our Framework}
\label{sec:method}

In \cref{sec:post_hoc_issue}, we investigated the potential of Annotation-based CBMs in addressing spurious correlations compared to CLIP-based CBMs. Nevertheless, building Annotation-based CBMs can be expensive due to the need for binary concept annotations.
In this section, we propose a framework of constructing Annotation-based CBMs \textit{with minimal human effort} using various foundation models.


Figure \ref{fig:framework} provides a comprehensive overview of our framework, which unfolds in three stages. 
First, we collect a concepts pool of visual concepts essential for classification, where spurious concepts are automatically filtered out.
Second, we use a MLLM to annotate the selected concepts for each image.
The optional third stage involves correcting imprecise concepts annotations by using chains of vision foundation models, which improves the reliability of the MLLM's annotations.
Using concept annotations obtained through three stages, we construct Annotation-based CBMs, denoted as \textit{LLaVA-based CBM}, effective in mitigating spurious correlations.
Detailed procedures are described in the following.

\subsection{Collecting Visual Attributes Unaffected by Spurious Correlation}
\label{sec:collect_va}
Constructing CBMs demands gathering a list of essential visual concepts, where concept quality significantly influences the classification model's capability. Particularly, mitigating spurious correlations requires careful concept selection to retain a list of concepts unrelated to such correlations. We propose a method outlined below, designed to collect diverse concepts while avoiding those linked with misleading correlations.

\noindent\textbf{Step 1: Use LLM to collect diverse visual concepts.} 
We leverage the capability of GPT-3~\cite{brown2020language} in identifying key concepts for target classes and collect diverse visual attributes for each classification task as in \cite{oikarinen2023label}.
Specifically, GPT-3 is prompted to provide \textit{important features}, \textit{superclass}, and \textit{things seen around} of each class via the OpenAI API. 
To further diversify the set of attributes, we additionally prompt GPT-3 with ``Provide a list of visual attributes to distinguish between a \textit{class 1} and \textit{class 2}''.
Then, we filter out concepts too similar to each other and to classes as in \cite{oikarinen2023label}.
We denote the collected concepts set as $S_1$.
However, we remark $S_1$ may include elements tied to spurious correlations. 

\noindent\textbf{Step 2: Use MLLM to detect potential spurious correlations.}\label{sec:step_2}
To identify potential spurious correlations within datasets, we propose two sub-steps using LLaVA (MLLM).  

\noindent \textbf{(i)} We collect descriptions for each image by prompting LLaVa with ``Describe the image in a sentence.''
We denote the description for input $i$ as $b_i$ and the collection of descriptions as $D=\{b_1, b_2, \dots, b_n\}$, where $n$ represents the number of training data.
Next, we leverage the in-context learning capability of GPT-3. We prompt GPT-3 with a few examples formatted as $(\text{description}, \text{keywords})$ to extract keywords of the descriptions and denote the collection of the keywords as $D'=\{k_1, k_2, \dots, k_n\}$, where ${k}_i = [\text{keyword}_1, \text{keyword}_2, \dots, \text{keyword}_o]$, and
$o$ varies with different inputs.
We append class labels to each $k_i$ to ensure each description contains the corresponding class.
\begin{figure}
\centering
\includegraphics[width=0.39\columnwidth]{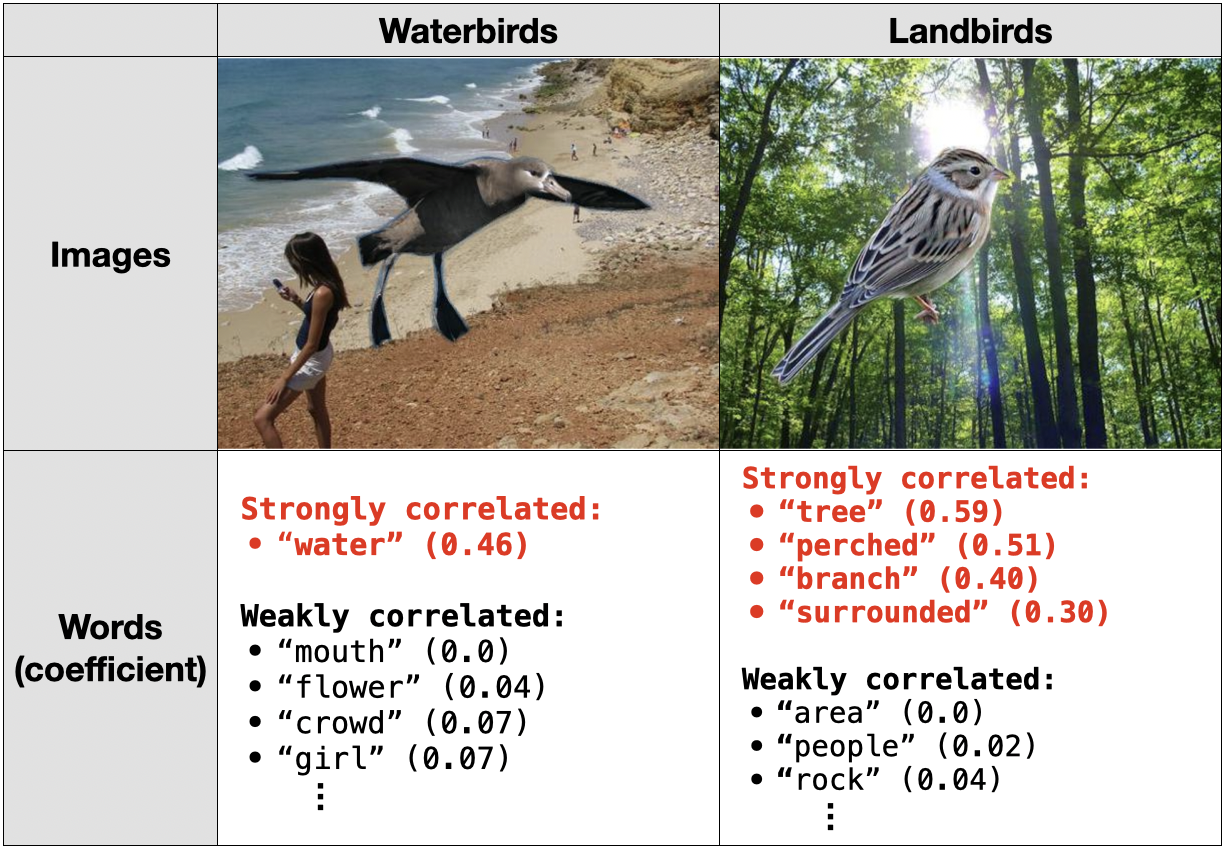}
    \caption{Potential spurious correlations detected on Waterbirds from Step 2. We enumerate strongly and weakly correlated words with coefficients, highlighting the strongly correlated words with red. The known spurious concepts are detected by our method. We set the threshold as 0.3 for coefficients to select highly correlated attributes.}
    \label{fig:spurious}
\end{figure}

\noindent \textbf{(ii)} We compute correlations between a class label and each keyword, and find the keywords showing high correlations with the label.
We start by creating a vocabulary set using $D'$. 
Assuming $N$ vocabularies, we vectorize each $k_i$ to represent the presence of each vocabulary, resulting in binary vectors ($\mathbf{v}$) of dimension $N$.
The collection of the vectors is denoted as $D''=\{\mathbf{v}_1,\mathbf{v}_2, \dots,\mathbf{v}_n\}$ , where $\mathbf{v}_i\in \{0,1\}^N$.
Each vocabulary is considered as a variable, yielding $n$ samples for the $N$ variables.
We measure correlations between the class labels and the $N-1$ vocabularies by computing the point biserial correlation coefficients, which range from -1 to 1.
By sorting keywords by their coefficients, we identify attributes highly correlated with the labels. 
We use a threshold to select high correlated attributes and denote them as $S_2$.
This method can detect spurious correlations. Figure \ref{fig:spurious} illustrates the process captures strong correlations between \textit{landbird} and the attributes related to the known spurious concepts, such as \textit{tree}.

\noindent\textbf{Step 3: Filtering concepts related to spurious correlations.}
We curate a concepts pool with $S_1$ and $S_2$ obtained from the previous process.
To filter out elements linked with spurious correlations, we exclude elements similar to those of $S_2$ from $S_1$.
Similarity between elements is computed based on an ensemble of CLIP text encoder and all-mpnet-base-v2 sentence embedding as in \cite{oikarinen2023label}.

The detailed procedure and the finalized list of attributes for each dataset are provided in \cref{app:prompt_attr}.

\subsection{Annotating the Concepts with LLaVA} \label{sec:stage_2}
In this stage, we generate binary concept annotations of the training data by querying a MLLM to decide if the curated concepts present in each image.
The prompts consist of both an image and text.
For example, we query LLaVA-v1.5-13B using prompts ``Does the object have \{\textit{attribute}\}?'' alongside the corresponding image. 
The text responses from LLaVA are then converted into binary numbers, which we use as concept annotations for each image.
Examples of prompts and responses are provided in \cref{app:prompts_framework}.
We acknowledge prompt engineering plays a crucial role in influencing the output of LLMs (MLLMs)~\cite{zhou2022large, white2023prompt, gu2023systematic}. 
In our work, we have minimally engaged in prompt engineering, leaving room for future optimization of prompts. 
We remark concept annotations used to require considerable human effort. Our framework effectively replaces the human effort with open-source MLLMs, like LLaVA, cutting down \textit{the cost to nearly zero}.

\noindent\textbf{Evaluating MLLM in concept annotation.}
LLaVA demonstrates its multimodal understanding capabilities through challenging benchmarks, such as coarse and fine-grained recognition tasks~\cite{liu2023improved}. 
However, annotations provided by LLaVA can often contain errors. 
To evaluate LLaVA's reliability as a proxy annotator, we examine the accuracy of the annotations.
For Waterbird, which already comes with 112 attributes and their labels from CUB, we can compare LLaVA's responses against the ground truth. 
It is worth noting, however, that the ground truth may not be completely accurate as they underwent post-processing to reduce annotation noise as detailed by \cite{koh2020concept}.
We discuss details in \cref{app:CUB_anno}.
\begin{wraptable}{l}{0.35\textwidth}
\caption{The evaluation of LLaVA's annotations against human annotations on Waterbirds.}
  \centering
  \begin{tabular}{@{}ccc@{}}
    \toprule
     Recall & Precision & F1 score\\
    \midrule
     0.92 & 0.21 & 0.31 \\
    \bottomrule
  \end{tabular}
  \label{tab:LLava_examine}
\end{wraptable}

Table \ref{tab:LLava_examine} presents the average recall, precision, and F1 score across all training images and 112 attributes.
The results indicate that LLaVA tends to respond positively to queries, resulting in noticeable false positives. 
Furthermore, we construct Annotation-based CBMs with LLaVA annotations and human annotations, respectively, and compare their performance. 
As indicated in \cref{tab:LLaVA_corr_effect}, the CBM with human annotations outperforms the CBM with LLaVA annotations, implying that the inferior annotation quality of LLaVA affects the performance of CBMs based on these annotations.

\begin{table}
\caption{Worst group accuracies and average accuracies on Waterbirds (we explain the metrics in \cref{sec:exp}) with different annotations. Human annotation achieves the highest accuracy \textit{at a considerable cost}.
}
  \centering
  \begin{tabular}{@{}lcc@{}}
    \toprule
    Annotator & Worst-group Acc. & Average Acc. \\
    \midrule
    Human  &  85.0\% & 98.4\% \\
    LLaVA  &  80.8\% & 83.8\% \\
    \bottomrule
  \end{tabular}
  \label{tab:LLaVA_corr_effect}
\end{table}

To reduce the accuracy gap between human anotations and LLaVA annotations without relying on human expertise, we introduce an optional automatic annotation refinement process in the following section.

\subsection{Automatic Annotation Refinement}
\label{sec:method_anno_ref}
Motivated by the observed imperfection of LLaVA annotations, we develop an optional refinement process for improved annotation quality. 
Devising adequate refinement requires us first to understand the imperfections of LLaVA. 
We examine randomly selected probing images and their corresponding response pairs to identify errors made by LLaVA.
Our inspection reveals LLaVA often confuses the image background with the main object. 
For instance, in classifying \textit{letter opener} and \textit{can opener} in ImageNet-Opener, \textit{wood material} is used as a visual concept (we will provide more details in \cref{sec:exp}).
We observe LLaVA makes frequent mistakes to answer if the object is made of wood material, especially when the object is placed on a \textit{wooden table}.  

\begin{wrapfigure}{r}{0.49\textwidth}
\centering
\includegraphics[width=0.4\columnwidth]{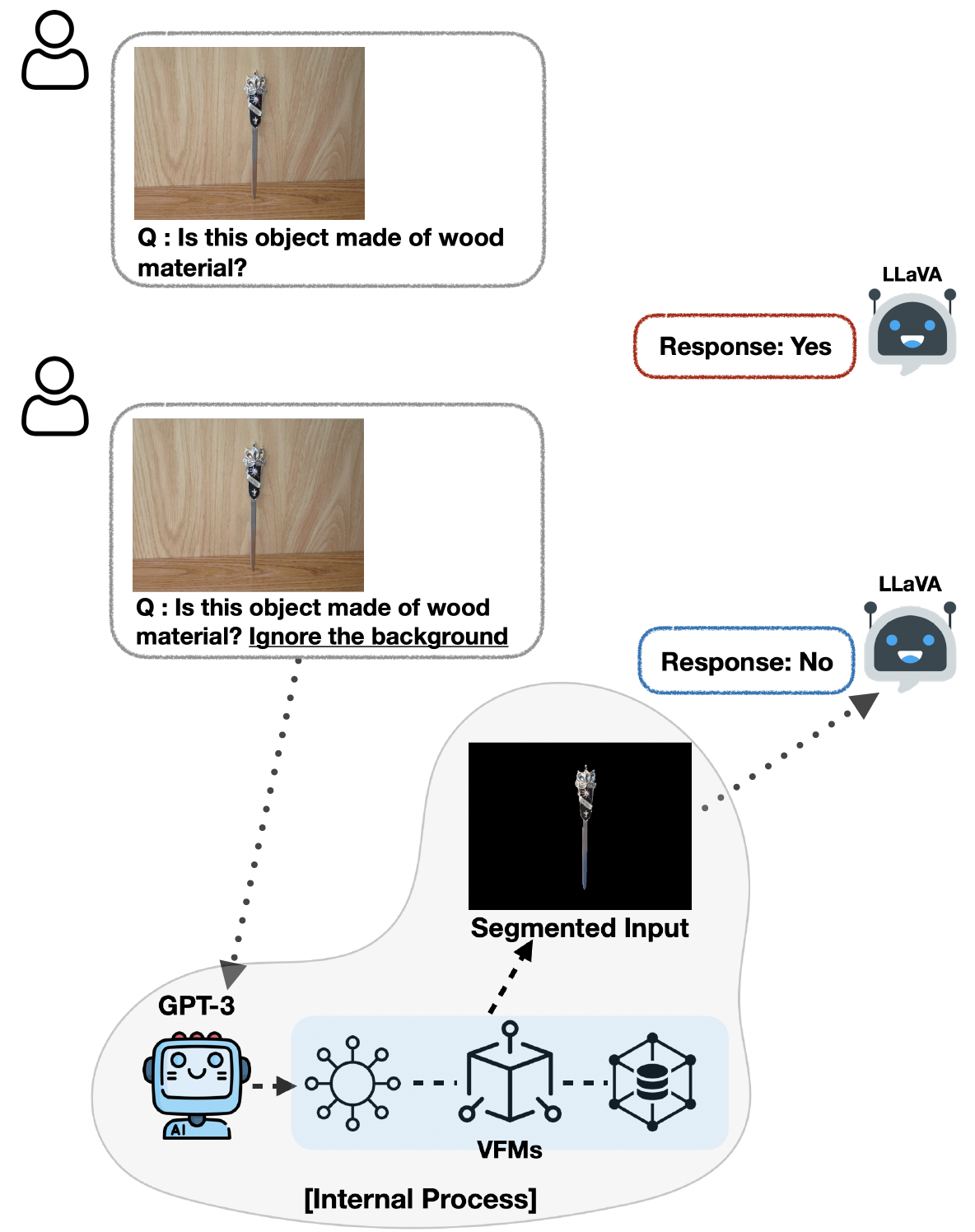}
    \caption{An illustration of annotation refinement process on ImageNet-Opener. The actual example demonstrates how the correction alters the response. The segmented input generated through a chain of VFMs is used as input to LLaVA. The internal process is not visible to human and executed automatically by GPT-3.}
    \label{fig:refinement}
\end{wrapfigure}

Once identifying LLaVA's errors, the next step entails correcting the errors with nearly no human effort, leveraging an LLM and VFMs. 
Recently, exploiting different VFMs to perform (sub-)tasks invoked by LLMs has been actively explored~\cite{chen2022visualgpt, chen2023llava, yang2023mm,gupta2023visual}.
LLM learns to utilize tools (e.g., VFMs) through in-context learning and activates these tools to accomplish a specific task. 
Drawing insights from probing LLaVA's errors, our task is set to rectify annotations by eliminating potentially confusing backgrounds in images before querying LLaVA.
To accomplish this, we employ Langchain~\cite{Chase_LangChain_2022} for implementation. We constitute a chain of tools composed of BLIP-2~\cite{li2023blip}, Grounding DINO~\cite{liu2023grounding} and SAM~\cite{kirillov2023segment}, leveraging the in-context learning capability of LLMs by providing specific examples that guide the activation of VFMs. 
For instance, the instructions, \textit{ignore the background} (\cref{fig:refinement}), trigger the chain of tools.
The LLM first activates the visual question answering model, BLIP-2, to identify the main object in the image. 
Next, it uses Grounding DINO to obtain the bounding box of the main object identified by BLIP-2. 
Then, the LLM applies SAM to segment the object associated with the bounding box.
Finally, the background-removed outputs from the series of tools are used to query LLaVA, making the annotations more accurate. 
Figure \ref{fig:refinement} shows an example of how the refinement corrects the annotations.
Note that our choice of VFMs is decided by the LLaVA's errors investigation. 
We presume VFMs to be interchangeable, and the selection of which VFMs to use can be guided by the potential errors identified in MLLM.

\section{Experiment}
\label{sec:exp}
We build \textit{LLaVA-based CBMs} based on the proposed framework and demonstrate the effectiveness of the CBMs in mitigating spurious correlations.

\subsection{Experimental Setups}
\noindent \textbf{Baselines.}
We compare \textit{LLaVA-based CBMs} with several methods including Empirical Risk Minimization (ERM) and CLIP-based CBMs (Label-free CBM~\cite{oikarinen2023label} and Post-hoc CBM~\cite{yuksekgonul2022post}).
Label-free CBM constructs a concept model using GPT-3 and CLIP.
On the other hand, Post-hoc CBM uses ConceptNet \cite{speer2017conceptnet} to generate concept sets and CLIP to obtain concept representations.
We provide the concept sets used in these models in \cref{app:clip_cbms_concepts}.
Furthermore, we compare ours with \cite{yang2023mitigating, wu2023discover}, which effectively mitigate spurious correlations and demonstrate state-of-the-art performances on datasets we use.
We include Group DRO \cite{sagawa2019distributionally} for comparison, known as an oracle, which requires group annotations.
Our method provides interpretability, thus our primary focus in comparisons is on other methods that also offer this feature.

\noindent \textbf{Datasets.} 
We assess our method on multiple challenging benchmarks that contain spurious correlations, as detailed below.
We provide the concepts we gather for each dataset in App.~\ref{app:visual_attr}. See App.~\ref{app:time_anno} for the annotation time details.

\noindent (i) ImageNet-Opener:
We use classes \textit{can opener} and \textit{letter opener} from a subset of ImageNet~\cite{deng2009imagenet} as suggested in \cite{yang2023mitigating}. 
In particular, \textit{can opener} usually appears together with \textit{can} in the training images, creating spurious correlations. 
When lacking \textit{can} in images, \textit{can opener} might be erroneously classified as \textit{letter opener}.

\noindent (ii) Metashifts:
Metashifts~\cite{liang2022metashift} provides many subsets of data corresponding to different contexts, naturally creating distribution shifts.
\cite{wu2023discover} proposes experiment setups for evaluating the model robustness to spurious correlations using Metashifts by creating disjoint spurious attributes for each class.
The training set includes two classes, \textit{cats} lying on \textit{soft/bed} and \textit{dogs} with \textit{bench/bike}. 
The testing set consists of \textit{cats} and \textit{dogs}, both with a \textit{shelf}.

\noindent (iii) Waterbirds:
Waterbirds~\cite{sagawa2019distributionally} comprises two classes: \textit{landbirds} and \textit{waterbirds}. 
In the training set, \textit{landbirds} and \textit{waterbirds} primarily appear on \textit{land} backgrounds and \textit{water} backgrounds, respectively, while the test set is balanced. 

\noindent \textbf{Evaluation Metrics.}
We assess the methods based on average accuracy and worst-group accuracy. 
For details on setting groups for each dataset, see App.~\ref{app:eval}.

\noindent \textbf{Model Architecture and Training Details.}
Both the ERMs and the concept models ($g$) of CBMs are built on ResNet-50~\cite{he2016deep}, pretrained on ImageNet. 
To build LLaVA-based CBMs, we train the concept model and the classifier independently. 
We provide more details on other hyperparameters in \cref{app:training_details}.
For fair comparison, we use the pre-trained CLIP encoders based on ResNet-50 for Label-free CBM and Post-hoc CBM. 
Label-free CBM and Post-hoc CBM results are reproduced using the official implementations\footnote{https://github.com/mertyg/post-hoc-cbm}\footnote{https://github.com/Trustworthy-ML-Lab/Label-free-CBM} with the identical configurations. 
We report results for all methods using early stopping based on worst-group validation accuracy, as per standard practice in prior works~\cite{sagawa2019distributionally}.
For Waterbirds, we use additional prompts to gather a richer concept pool. We provide more details on the prompts used in Waterbirds in \cref{app:prompts_framework}.

\subsection{Results}
\begin{table*}
\caption{Worst group and average accuracies on Metashifts and ImageNet-Opener. 
\cite{wu2023discover} identifies spurious concepts but lacks interpretability for predictions.
  In ImageNet-Opener, refined annotations on two attributes based on our proposed method are used.
  Results from the original papers are marked with $^*$. The ERM's and Group DRO's results on Metashift and Imagenet-Opener are from\cite{wu2023discover} and \cite{yang2023mitigating}, respectively.}
  \centering
  \begin{tabular}{@{}l|c|cc|cc@{}}
    \toprule
    & &\multicolumn{2}{c}{\textit{Metashifts}}& \multicolumn{2}{c}{\textit{ImageNet-Opener}} \\
    Method &  Interpretability & Worst Acc.  & Avg. Acc.  &Worst  Acc. & Avg.  Acc. \\
    \midrule
    ERM$^*$ &  \xmark &62.1\% & 72.9\%&68.0\% & 80.1\% \\
    DISC~\cite{wu2023discover}$^*$ &  $\triangle$&  73.5\% & 75.5\% &$\cdot$ &$\cdot$\\
    \cite{yang2023mitigating}$^*$  & \xmark&  $\cdot$ &$\cdot$ & 68.0\% & 73.9\%\\
    Label-free CBM \cite{oikarinen2023label}  & \cmark & 64.1\% & 74.1\% & 82.5\% & 78.0\%\\
    Post-hoc CBM \cite{yuksekgonul2022post}  & \cmark & 71.4\% & 80.4\% & 45.0\% & 48.0\%\\
    LLaVA-based CBM & \cmark& \bf{78.0\%} & \bf{80.0\%} & \bf{90.0\%} & \bf{86.1\%}\\
    \midrule
    Group DRO~\cite{sagawa2019distributionally}{$^*$}& \xmark& 66.0\% & 73.6\% & 76.0\% & 78.4\% \\
    \bottomrule
  \end{tabular}
  \label{tab:metashift_imagenet_result}
\end{table*}
\noindent \textbf{Quantitative Results.}
Table \ref{tab:metashift_imagenet_result} and Table \ref{tab:waterbird_result} report the average and the worst-group accuracies of different methods across various datasets.
LLaVA-based CBMs outperform other baseline models in most cases.
ERM presents a significantly lower worst-group accuracy than the average accuracy, suggesting its strong reliance on spurious correlations.
Although CLIP-based CBMs may narrow the gap between the worst-group and the average accuracy on Metashifts and ImageNet-Opener, the improvements remain limited.
Moreover, when compared to the state-of-the-art methods, such as \cite{yang2023mitigating} for ImageNet-Opener, DISC~\cite{wu2023discover} for Metashift, and Group DRO~\cite{sagawa2019distributionally} as an oracle for both, our method exhibits notably superior performance.
Although LLaVA-based CBM may not outperform Human Annotation CBM or Group DRO in \cref{tab:waterbird_result}, it is worth noting that these methods rely on human efforts, which can be \textit{significantly costly}.
Moreover, LLaVA-based CBMs can offer interpretability, which is missing in ERM or Group DRO. Additionally, we provide the results of using LLaVA directly for label prediction in App.~\ref{app:llava_direct}.

\begin{table}
\caption{Worst group accuacies and average accuracies on Waterbirds. 
1st/2nd best accuracies are marked with bolded/underlined. 
\textit{Post-hoc CBM} uses 112 CUB attributes in~\cite{koh2020concept} and \textit{Label-free CBM} uses 370 CUB attributes in~\cite{oikarinen2023label}.
\textit{Human Anno. CBM} denotes Annotation-based CBM with 112 CUB attributes and their labels in~\cite{koh2020concept}.
}
  \centering
  \begin{tabular}{@{}lcc@{}}
    \toprule
    Method &  Worst-group Acc. & Average Acc.  \\
    \midrule
    ERM & \underline{72.6\%} & \bf{97.3\%} \\
    Label-free CBM~\cite{oikarinen2023label}  &  55.0\% & 82.8\%\\
    Post-hoc CBM~\cite{yuksekgonul2022post} &   31.6\% & 91.7\%\\
    LLaVA-based CBM &  \bf{83.2\%} & \underline{94.2\%} \\
    \midrule
    Human Anno. CBM (CUB) &  85.0\% & {98.4\%} \\
    Group DRO~\cite{sagawa2019distributionally} & {91.4}\% & 93.5\% \\
    \bottomrule
  \end{tabular}
  \label{tab:waterbird_result}
\end{table}
Particularly noteworthy is the superiority of LLaVA-based CBM over CLIP-based CBMs~\cite{yuksekgonul2022post, oikarinen2023label} across various datasets.
Note that CLIP-based CBMs rely on concepts from CUB, which focus solely on spurious-free bird appearance concepts as clarified in \cref{sec:post_hoc_issue}, and hinder the model generalization to novel applications.
Meanwhile, our approach is not dependent on CUB, making it applicable to a wide range of datasets.
Our superior performance can be attributed to two potential factors.
First, as discussed in \cref{sec:post_hoc_issue}, we posit biases inherent in CLIP can negatively affect CBMs built on it.
Table \ref{tab:waterbird_result_add} further supports this claim.
Despite utilizing identical concept sets that exclude spurious elements, CLIP-based CBMs exhibit a significant drop in worst-group accuracy, a shortcoming considerably mitigated in LLaVA-based CBM.
Second, our strategy to gather visual attributes unaffected by spurious correlations, as proposed in ~\cref{sec:collect_va}, proves effective, ensuring that the collected concept sets are unlikely to include concepts relevant to spurious correlations. 
Refer to App.~\ref{app:clip_cbms_concepts} for the used concept sets in CLIP-based CBMs across datasets.
\begin{table}
\caption{Worst group and average accuracies on Waterbirds.  
\textit{Post-hoc CBM} and \textit{LLaVA-based CBM (CUB)} use 112 CUB attributes in~\cite{koh2020concept}.
\textit{Label-free CBM} and \textit{LLaVA-based CBM (Label-free)} use 370 CUB attributes in~\cite{oikarinen2023label}.
}
  \centering
  \begin{tabular}{@{}lcc@{}}
    \toprule
    Method &  Worst-group Acc. & Average Acc.  \\
    \midrule 
    Post-hoc CBM~\cite{yuksekgonul2022post} &   31.6\% & 91.7\%\\
    LLaVA-based CBM (CUB) & \bf{80.8\%}   &  83.8\% \\
    \midrule
    Label-free CBM~\cite{oikarinen2023label}  &  55.0\% & 82.8\%\\
    LLaVA-based CBM (Label-free) &  {80.5\%} & \bf{94.1\%}  \\
    \bottomrule
  \end{tabular}
  \label{tab:waterbird_result_add}
\end{table}

\noindent \textbf{Spurious Correlations Detection.}
We showcase concept pools collected using the procedure outlined in \cref{sec:collect_va}.
For ImageNet-Opener, we collect 47 visual concepts by prompting GPT-3 as Step 1.
Step 2 identifies the words ``can'', ``person'', and ``opener'' to be highly correlated to the class labels. The known spurious concept ``can'' is detected in this step.
We exclude five visual concepts identified as similar to the detected spurious concepts in Step 3 from the initial 47, leaving us with 42 visual attributes.

\begin{wrapfigure}{r}{0.47\textwidth}
\centering
\includegraphics[width=0.4\columnwidth]{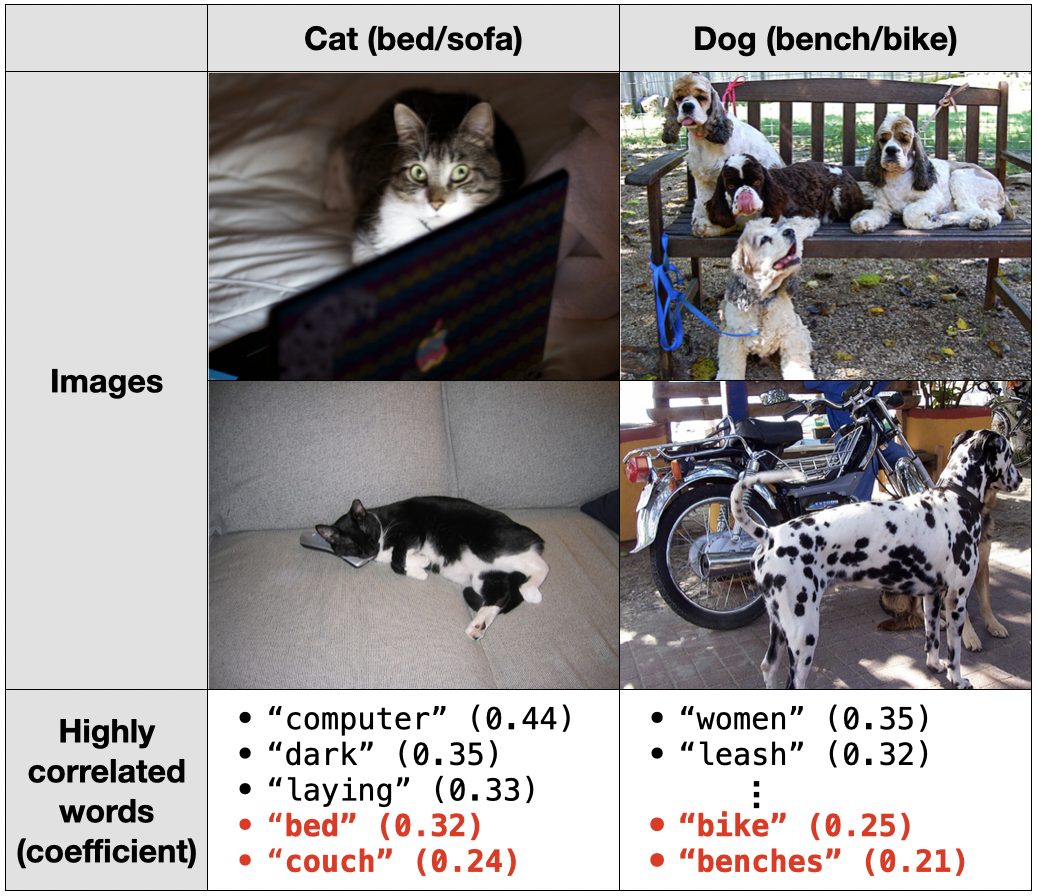}
    \caption{Potential spurious correlations detection on Metashifts. All known spurious concepts (bed, sofa, bench, and bike colored as red) are detected by our method. We set the threshold as 0.2 for correlation coefficients to select correlated words. See App.~\ref{all:threshold} for details on varying thresholds.}
    \label{fig:spurious_metashift}
\end{wrapfigure}
In Metashifts, among the 28 visual concepts collected from Step 1, Step 2 detects 16 words including ``benches'', ``frisbee'', ``bike'', ``leash'', ``couch'', ``bed'' and ``computer'' as potential spurious concepts. 
Interestingly, this set contains not only the known spurious concepts in Metashifts such as ``bed'', ``couch'', ``bike'' and ``bench'', but also other potential spurious concepts such as ``leash'', ``compute'' and ``frisbee''. 
We filter out 3 keywords  (``leash'', ``a cat bed'', ``a cat toy'') based on similarity from our initial concepts pool and obtain 24 visual concepts free from biases. 
We show the Step 2 examples of Metashifts in \cref{fig:spurious_metashift}.
We provide examples in Waterbirds in App.~\ref{app:visual_attr}.

\noindent \textbf{The Effect of Automatic Annotation Refinement.}
We use ImageNet-Opener to assess the effectiveness of the annotation refinement process introduced in~\cref{sec:method_anno_ref}.
As discussed in \cref{sec:method_anno_ref}, one of the common errors made by LLaVA is incorrectly identifying objects as being made of wood, when the opener is placed on a wooden table. 
To mitigate such errors, we refine annotations of two specific concepts, ``wood material'' and ``knife-like shape'', through the proposed refinement process using either LLM or a chain of vision foundation models. 

In our qualitative evaluation of LLaVA's responses, we focus on changes in LLaVA's responses before and after the refinement.
For ``wood material'',  21\% of \textit{can opener} responses identified as made of wood initially, reduced to 3\% after refinement, aligning better with common knowledge. On the other hand,  positive answers to ``wood material'' decreased from 41\% to 11\% in \textit{letter opener}.  
Regarding ``knife-like shape'', positive identifications in \textit{can opener} dropped from 90\% to 86\%, while for \textit{letter opener}, it marginally reduced from 98.5\% to 98.0\%, reflecting the expected relevance of ``knife-like shape'' to \textit{letter opener}.

Based on the refined annotation, we examine the impact of the refinement process on classification performance.

\begin{wraptable}{r}{0.5\textwidth}
\caption{Worst-group and average accuracy on Imagenet-Opener. We refined annotations on two attributes (\textit{wood material} and \textit{knife-like shape}). The refinement enhances the accuracy.}
  \centering
  \begin{tabular}{@{}lcc@{}}
    \toprule
     & Worst. Acc. & Avg. Acc.  \\
    \midrule
    without refinement &  80.0\% & 82.4\%\\
    with refinement & \bf{90.0\%}   & \bf{86.1\%} \\
    \bottomrule
  \end{tabular}
\label{tab:imagenet_result_refine}
\end{wraptable}
Table \ref{tab:imagenet_result_refine} shows that the refinement positively impacts accuracy gains compared to the CBM without annotation refinement.
The result highlights the effectiveness of the annotation refinement in error correction and the subsequent improvement in the accuracy of LLaVA-based CBMs.

\noindent \textbf{Interpretability of LLaVA-based CBMs.}
We examine how the models derive their predictions, which is one of the advantages of employing CBMs.
For ImageNet-Opener, we identify the most common attributes associated with correctly predicted test samples. Specifically, all correctly predicted \textit{letter opener} samples share 19 positive attributes, while correct \textit{can opener} samples consistently predict 21 specific attributes positively. There are 17 overlapping attributes for both classes. 
We focus on the non-overlapping attributes for each class and observe these concepts align well with common visual attributes associated with each class, as following:
\begin{itemize}
    \item \textit{Letter opener}:  \{``sharp blade'', ``an ornamental handle'' \}
    \item \textit{Can opener}:  \{``a round wheel blade'', ``a bottle opener'', ``a lever for opening cans'', ``a counter''\}
\end{itemize}
We further explore incorrect test samples to understand why they make wrong predictions. For example, in some incorrect samples of \textit{letter opener}, the model fails to identify the distinct concept ``an ornamental handle'' and tends to positively predict all concepts belonging to \textit{can opener}. 
On the other hand, Label-free CBM's top 2 weights to predict \textit{can opener} are assigned to the concepts, ``a small, handheld tool'' and ``a can''. In the case of \textit{letter opener}, they are ``a long, thin blade'' and ``a pen''. 
Thus, the image of \textit{can opener} lacking ``a can'' tends to be misclassified, while correct prediction occurs if ``a can'' is present.
The examples are illustrated in \cref{fig:label_free_ex}.
The results suggest that LLaVA-based CBMs are beneficial for mitigating spurious correlations, unlike Label-free CBMs.

\section{Conclusion}

In this paper, we explored mitigating spurious correlations using CBMs developed with minimal human effort by integrating multiple foundation models.
We introduced a framework that collects visual concepts that are essential but not affected by spurious correlation and performs concept annotation for each image to build CBMs, all leveraging the capabilities of MLLM and LLM.
Furthermore, we suggest an optional refinement method for the annotations to further improve the reliability of the annotations by an MLLM. 
We demonstrate the effectiveness of our approach in tackling spurious correlations on diverse real-world challenges. 

\section*{Acknowledgements}
This work was supported by NIH (1U01CA269192).
%
%
\bibliographystyle{splncs04}
\bibliography{main}



\clearpage
\setcounter{page}{1}
\appendix
\setcounter{table}{0}
\setcounter{equation}{0}
\setcounter{figure}{0}
\renewcommand{\thetable}{\Alph{table}}
\renewcommand{\thefigure}{\Alph{figure}}

\section{Prompts and Visual Concepts} \label{app:prompt_attr}
\subsection{Prompts}
\label{app:prompts_framework}
\noindent\textbf{Stage 1 in \cref{sec:collect_va}.}
In Step 1, we prompt GPT-3 to get \textit{important features}, \textit{superclass}, \textit{things seen around} \cite{oikarinen2023label} and \textit{distinguished visual features} of each class. 
For the queries related to \textit{important features}, \textit{superclass}, and \textit{things seen around}, we use the in-context learning examples as described in \cite{oikarinen2023label}.
The prompts used to query GPT-3 are the following:
\begin{itemize}
    \item \textit{important features} - List the most important features for recognizing something as a \{class\}
    \item \textit{superclass} - Give superclasses for the word \{class\}
    \item \textit{things seen around} - List the things most commonly seen around a \{class\}
    \item \textit{distinguished visual features} - Provide a list of visual features to distinguish between a \{class 1\} and \{class 2\}
\end{itemize}

For Waterbirds, we use different prompts to collect a more comprehensive concept set.
We first obtain \textit{subclass} of each class. 
This yields a total of 171 distinct bird species initially, and through a subsequent process using the same \textit{subclass} prompts, we obtain a total of 1852 bird species. 
Then, we use the prompts suggested above for \textit{important features}, \textit{superclass} and \textit{things seen around}.

In Step 2, we obtain image descriptions by prompting the LLaVA with ``Describe the image in a sentence'' and the corresponding image.
Then, we extract keywords from the list of descriptions by querying GPT-3.
We leverage the in-context learning capabilities of GPT-3 to extract keywords from the descriptions. The prompts used to query GPT-3 for each dataset are detailed in \cref{tab:prompts}.
We use nltk\footnote{https://www.nltk.org} tokenizer to tokenize the collection of keywords and remove prepositions. 
We compute the point biserial correlation coefficients to detect potential spurious features. To prevent the correlation coefficients from being distributed among similar words, we combine the similar words into one based on similarity (using CLIP text encoder and all-mpnet-base-v2 sentence embedding) and add the correlation coefficients of each word.

\begin{table*}
\caption{The prompts used to extract keywords from the image descriptions in Step 2.}
  \centering
  \resizebox{\textwidth}{!}{\begin{tabular}{@{}l|l@{}}
    \toprule
    Datasets &  Prompts  \\
    \midrule
    
    & Instruction : Extract concepts from the sentence such as the examples below. \\
    &The first example, \\
    & \qquad sentence : ``A can opener is sitting on top of a can of food.'',\\
    & \qquad concepts : ``can opener, a can of food'' \\
    Imagenet-Opener& The second example, \\
    & \qquad sentence : ``A knife with a wooden handle is sitting on a green plate.'', \\
    & \qquad concepts : ``knife with a wooden handle, green plate'' \\
    & The third example, \\
    & \qquad sentence : ``A blue wrench with a handle is shown in a close-up view.'', \\
    & \qquad concepts : ``blue wrench with a handle'' \\
    \midrule
    &  Instruction : Extract concepts from the sentence such as the examples below. \\
    &The first example, \\
    & \qquad sentence : ``An orange tabby cat is laying on a bed next to a laptop computer.'',\\
    & \qquad concepts : ``tabby can, laying on a bed, laptop computer'' \\
    Metashifts & The second example, \\
    & \qquad sentence : ``A cat is laying on a blanket and eating food from a fork.'', \\
    & \qquad concepts : ``cat, laying on a blanket, eating food'' \\
    & The third example, \\
    & \qquad sentence : ``A cat is sleeping on a couch, with its head resting on a pillow.'', \\
    & \qquad concepts : ``cat, sleeping on a couch, resting on a pillow'' \\
    \midrule
     & Instruction : Extract concepts from the sentence such as the examples below. \\
    &The first example, \\
    & \qquad sentence : ``The image features a seagull flying over the ocean, with its wings spread wide open as it soars \\
    & \qquad \qquad \qquad \quad through the sky.'',\\
    & \qquad concepts : ``seagull flying over the ocean, wings spread wide open.'' \\
    Waterbirds& The second example, \\
    & \qquad sentence : ``The image shows a seagull floating on water, with its wings spread out, and its body resting \\
    & \qquad \qquad \qquad \quad on the surface.'', \\
    & \qquad concepts : ``seagull floating on water, wings spread out, body resting on the surface.'' \\
    & The third example, \\
    & \qquad sentence : ``The image features a bird perched on a tall bamboo plant, with a blue and black color scheme.'', \\
    & \qquad concepts : `` perched on a tall bamboo plant, with a blue and black color scheme'' \\
    \bottomrule
  \end{tabular}}
  \label{tab:prompts}
\end{table*}

\textbf{Stage 2 in \cref{sec:stage_2}}
Table \ref{tab:llava_prompt_imagenet} and Table \ref{tab:llava_prompt_metashift} illustrate the prompts for querying LLaVA to annotate the concepts of each image for ImageNet-Opener and Metashifts. 
Note that the concepts are obtained from Stage 1.
For Waterbirds, we use \textit{``Does the object have} \{\textit{attribute}\}?'' for all attributes.

\begin{table*}
\caption{The prompts used for LLaVA to annotate the concepts of each image in ImageNet-Opener.}
  \centering
  \begin{tabular}{@{}l@{}}
    \toprule
  Prompts  \\
    \midrule
     Does the object have a bulkier shape with gears and handles? \\
     Does the object have a round cutting wheel shape? \\
     Is the object made of metal material?\\
     Is the object made of plastic material?\\
     Is the object made of rubber material?\\
     Does the object have a round wheel blade?\\
     Does the object have a blade not overly sharp to touch?\\
     Does the object have a handle for better grip?\\
     Does the object have a bottle opener?\\
     Does the object have a lid lifter?\\
     Does the object have a sleek shape?\\
     Does the object have a streamlined shape?\\
     Does the object have a knife-like shape? \\
     Does the object have a delicate size?\\
     Is the object made of wood material?\\
     Is the object made of ornamental material?\\
     Does the object have a solid piece?\\
     Does the object have a flat blade?\\
     Does the object have a sharp blade?\\
     Does the object have a slim handle?\\
     Does the object have an ornamental handle?\\
     Does the object have a blade for cutting open cans?\\
     Does the object have a blunt end?\\
     Does the object have a chair?\\
     Does the object have a comfortable grip?\\
     Does the object have a computer?\\
     Does the object have a counter?\\
     Does the object have a cupboard?\\
     Does the object have a desk?\\
     Does the object have a durable construction?\\
     Does the object have a fridge?\\
     Does the object have a kitchen?\\
     Does the object have a lever for opening cans?\\
     Does the object have a long, thin blade?\\
     Does the object have a mailbox?\\
     Does the object have a pen?\\
     Does the object have a pointed end?\\
     Does the object have a printer?\\
     Does the object have a small handle?\\
     Does the object have a stamp?\\
     Does the object have envelopes?\\
     Does the object have a paper?\\
     \bottomrule
  \end{tabular}
  \label{tab:llava_prompt_imagenet}
\end{table*}

\begin{table*}
\caption{The prompts used for LLaVA to annotate the concepts of each image in Metashifts.}
  \centering
  \begin{tabular}{@{}l@{}}
    \toprule
  Prompts  \\
    \midrule
     Does the object have a collar? \\
     Does the object have a dog tag? \\
     Does the object have a food bowl?\\
     Does the object have a litter box?\\
     Does the object have a person?\\
     Does the object have a scratching post?\\
     Does the object have a toy?\\
     Does the object have floppy ears?\\
     Does the object have a semi-erect ears?\\
     Does the object have a prominent snout with more separated nostrils?\\
     Does the object have a curly tail?\\
     Does the object have a straight tail?\\
     Does the object have a fluffy tail?\\
     Does the object have a sleek tail?\\
     Does the object have a smooth fur?\\
     Does the object have a curly fur?\\
     Does the object have a wirly fur?\\
     Does the object have round paws?\\
     Does the object have erect ears?\\
     Does the object have a short snout?\\
     Does the object have a flat snout with closely set nostrils?\\
     Does the object have a slender tail?\\
     Does the object have a flexible tail?\\
     Does the object have oval-shaped paws with retractable claws?\\
     \bottomrule
  \end{tabular}
  \label{tab:llava_prompt_metashift}
\end{table*}

\subsection{Collected Visual Attributes in \cref{sec:collect_va}} \label{app:visual_attr}
We collect the visual concepts relevant to each class from Step 1 and they are described in \cref{tab:step_1_details}.
We refrain from enumerating the visual concepts of Waterbirds due to their large number (\textbf{444} attributes).
We detect potential spurious features in Step 2 as detailed in \cref{tab:step_2_details}.
In Step 3, we finalize the visual concepts for constructing CBMs by filtering out potentially spurious features from the relevant concepts based on their similarity.
The visual concepts used for constructing CBMs in each dataset are described in \cref{tab:final_va}.
We refrain from enumerating the visual concepts of Waterbirds due to their large number (\textbf{428} attributes).
The concepts including ``a palm tree'', ``a perching posture'', ``a green body color'' and ``a beach or ocean scene'' are filtered out in Step 3.

\begin{table*}
\caption{The collected relevant visual concepts in Step 1.}
  \centering
  \resizebox{\textwidth}{!}{\begin{tabular}{@{}l|l@{}}
    \toprule
    Datasets &  Prompts  \\
    \midrule
    & ``a blade for cutting open cans'', ``a blunt end'', ``a can'', ``a chair'', ``a comfortable grip'',\\
    & ``a computer'', ``a counter'', ``a cupboard'', ``a desk'', ``a durable construction'', ``a fridge'' \\
    & ``a kitchen'', ``a lever for opening cans'', `a long, thin blade'', ``a mailbox'', \\
    & ``opener'', ``a pen'', ``a person'', ``a pointed end'', ``a printer'', \\
    Imagenet-Opener& ``a stamp'', ``a table'', `envelopes'', ``paper'', ``knife'', ``a bulkier shape with gears and handles'', \\
    & ``a round cutting wheel shape'', ``metal material'', ``plastic material'', ``rubber material'' , ``a round wheel blade'',\\
    &``a blade not overly sharp to touch'', ``a handle for better grip'',``a bottle opener'', ``a lid lifter'', ``a sleek shape'',\\
    &``a streamlined shape'', ``a knife-like shape'', ``delicate size'', ``wood material'',  ``ornamental material'',\\
    & ``solid piece'', ``a flat blade'', ``a sharp blade'', ``a slim handle'', ``an ornamental handle'' \\
    \midrule
    &  ``a collar'', ``a dog tag'', ``a food bowl'', ``a litter box'', ``a person'', ``a scratching post'', ``a toy'',\\
    &``leash'', ``a cat bed'', ``a cat toy'', ``floppy ear'',``semi-erect ears'', ``prominent snout with more separated nostrils'',\\
    Metashifts &``a curly tail'',``a straight tail'',
    ``a fluffy tail'', ``a sleek tail'', ``a smooth fur'', ``a curly fur'', ``a wirly fur'',\\
    &``round paws'', ``erect ears'', ``a short snout'', ``a flat snout with closely set nostrils'',\\
    &``a slender tail'', ``a flexible tail'', ``oval-shaped paws with retractable claws'', ``mammal''\\
    \bottomrule
  \end{tabular}}
  \label{tab:step_1_details}
\end{table*}

\begin{table*}
\caption{The potential spurious features and their coefficients detected in Step 2. The threshold for the coefficient is set to 0.2.}
  \centering
  \begin{tabular}{@{}l|l@{}}
    \toprule
    Datasets &  Prompts  \\
    \midrule
    &\textit{Can Opener} - can (0.39) \\
    & \qquad \qquad \quad \;\; opener (0.24) \\
    & \qquad \qquad \quad \;\; person (0.22) \\
    Imagenet-Opener& \textit{Letter Opener} - knife (0.52) \\
    & \qquad \qquad \quad \;\;\;\; wood (0.28) \\
    & \qquad \qquad \quad \;\;\;\; gold (0.27) \\
    & \qquad \qquad \quad \;\;\;\; design (0.26) \\
    & \qquad \qquad \quad \;\;\;\; surface (0.20) \\
    \midrule
    &  \textit{Cat} - kitty (0.96) \\
    &  \quad \;\;\;\; computer (0.44) \\
    & \quad \;\;\;\; dark (0.35) \\
    & \quad \;\;\;\; laying (0.33) \\
    & \quad \;\;\;\; bed (0.32) \\
    & \quad \;\;\;\; couch (0.24) \\
    Metashifts & \textit{Dog} - puppy (0.97) \\
    & \quad \;\;\;\; women (0.35) \\
    & \quad \;\;\;\; leash (0.32) \\
    & \quad \;\;\;\; walking (0.30) \\
    & \quad \;\;\;\; small (0.27) \\
    & \quad \;\;\;\; bike (0.25) \\
    & \quad \;\;\;\; frisbee (0.23) \\
    & \quad \;\;\;\; catching (0.20) \\
    & \quad \;\;\;\; benches (0.20) \\
    & \quad \;\;\;\; jumping (0.20) \\
    \midrule
     &\textit{Waterbird} - water (0.46) \\
    & \qquad \qquad \quad body (0.27) \\
    & \qquad \qquad \quad  beach (0.27) \\
    & \qquad \qquad \quad  bird (0.27) \\
    & \qquad \qquad \quad  background (0.27) \\
    & \qquad \qquad \quad  ship (0.26) \\
    & \qquad \qquad \quad  seagull (0.24) \\
    & \qquad \qquad \quad  large (0.24) \\
    & \qquad \qquad \quad  duck (0.21) \\
    & \qquad \qquad \quad  standing (0.2) \\
    Waterbirds& \textit{Landbird} - tree (0.58) \\
    & \qquad \qquad \; perched (0.50) \\
    & \qquad \qquad \;  surrounded (0.39) \\
    & \qquad \qquad \;  bamboo (0.27) \\
    & \qquad \qquad \;  green (0.24) \\
    & \qquad \qquad \;  plant (0.24) \\
    & \qquad \qquad \;  small (0.23) \\
    & \qquad \qquad \;  path (0.23) \\
    \bottomrule
  \end{tabular}
  \label{tab:step_2_details}
\end{table*}

\begin{table*}
\caption{The visual concepts used for constructing CBMs in each dataset.}
  \centering
  \begin{tabular}{@{}l|l@{}}
    \toprule
    Datasets &  Prompts  \\
    \midrule
    &``a blade for cutting open cans'', \\
    &``a chair'', ``a comfortable grip'',\\
    & ``a computer'', ``a counter'', ``a cupboard'', \\
    & ``a desk'', ``a durable construction'',  \\
    & ``a kitchen'', ``a lever for opening cans'', \\
    &`a long, thin blade'', ``a mailbox'', \\
    &  ``a pen'', ``a pointed end'', ``a printer'', \\
    Imagenet-Opener& ``a stamp'', `envelopes'', ``paper'',\\
    &``a bulkier shape with gears and handles'',\\
    &``a round cutting wheel shape'',\\
    &``metal material'', ``plastic material'',\\
    &``rubber material'' , ``a round wheel blade'',\\
    &``a blade not overly sharp to touch'',\\
    &``a handle for better grip'',``a bottle opener'',\\
    &``a lid lifter'', ``a sleek shape'', ``a fridge'',\\
    &``a streamlined shape'', ``a blunt end'', \\
    &``a knife-like shape'', ``delicate size'',\\
    &``wood material'',  ``ornamental material'',\\
    & ``solid piece'', ``a flat blade'', \\
    &``a slim handle'', ``an ornamental handle'', \\
    & ``a sharp blade''\\
    \midrule
    &   ``a collar'', ``a dog tag'', \\
    &``a food bowl'', ``a litter box'', ``a person'',\\
    &``a scratching post'', ``a toy'',\\
    &``floppy ear'',``semi-erect ears'', \\
    &``prominent snout with more separated nostrils'',\\
    Metashifts &``a curly tail'',``a straight tail'',\\
    &``a fluffy tail'', ``a sleek tail'',\\
    &``a smooth fur'', ``a curly fur'', ``a wirly fur'',\\
    &``round paws'', ``erect ears'', ``a short snout'',\\
    &``a flat snout with closely set nostrils'',\\
    &``a slender tail'', ``a flexible tail'', \\
    &``oval-shaped paws with retractable claws'' \\
    \bottomrule
  \end{tabular}
  \label{tab:final_va}
\end{table*}

\section{Concept Annotations of CUB Post-processing} \label{app:CUB_anno}
The CUB dataset~\cite{wah2011caltech} comprises 11,788 images of birds from 200 species, with each image additionally annotated with binary concepts that denote specific bird attributes, such as wing color and beak shape.
\cite{koh2020concept} refine the concept annotations of CUB to build CBMs.
To enhance the accuracy and coherence of annotations, which is compromised due to contributions from multiple crowdworkers (not a bird expert), they implement a majority voting to consolidate instance-level concept annotations into class-level concepts: \eg, if more than 50\% of crows have black wings in the data, then they set all crows to have black wings. This makes the approximation that all birds of the same species in the training data should share the same concept annotations. After majority voting, they further filter out sparse concepts, retaining only concepts that are present after majority voting in at least 10 classes. This process results in a refined list of 112 concept annotations.

\section{Experiment Details}
\subsection{Training Details}
\label{app:training_details}
Annotation-based CBMs comprise two main components: a concept model and a classifier.
The concept models consist of the ResNet-50 architecture for feature extraction and multiple single-layer classifiers built on top of the feature extractor. 
The concept models are trained to predict individual concepts with the binary cross-entropy loss where each individual concept prediction task is weighted by the ratio of class imbalance for that individual concept, as suggested in \cite{koh2020concept}.
We use a batch size of 64, a learning rate of 0.01, and SGD with momentum of 0.9 as the optimizer to train the concept models.
The classifiers of annotation-based CBMs are constructed with a small four-layer MLP.
We train classifiers with a learning rate of 0.001, a batch size of 64, and SGD with momentum of 0.9 as the optimizer. 

\subsection{The Time for Data Annotations}
\label{app:time_anno}

\begin{table}[h]
\caption{The time for data annotation in Stage 2 (on two GPUs).}
    \centering
    \begin{tabular}{c|c|c|c|c}
    \toprule
    &\# of concepts&  \# of images & time (s) for 1 image &Total time (h) \\
    \midrule
        Waterbirds&  428 & 5,914 & 180 &295 \\
         Metashift& 24 & 1,105&10&3\\
         ImageNet-Opener& 42 & 2,286&18&11\\
        \bottomrule
    \end{tabular}
    \label{tab:data_time}
\end{table}
\noindent In Tab.~\ref{tab:data_time}, we quantify the annotation speed on two NV3090 GPUs. Multi-GPU parallelization is used for expedite the Waterbirds annotation. \textit{We emphasize the substantially reduced cost compared to human annotation}. For instance, each of the 312 concepts in CUB is annotated by a crowdworker with additional postprocessing (See App.~\ref{app:CUB_anno}). 

\subsection{Evaluation Metrics}
\label{app:eval}
We assess the methods based on average accuracy and worst-group accuracy. 
For Waterbirds, the worst-group accuracy denotes the lowest accuracy across groups, defined by the combination of spurious attributes and the classes, and the average accuracy is the adjusted classification accuracy averaged over groups weighted according to their sizes in the training data, as in~\cite{sagawa2019distributionally}. 
For ImageNet-Opener, the adjusted classification accuracy is used as average accuracy and the worst-performing group is identified as \textit{can opener} images without the presence of \textit{can}, which is considered as the spurious concepts, as in \cite{yang2023mitigating}.
For Metashifts, the worst group accuracy indicates the accuracy on class \textit{dog}, and the average accuracy denotes the accuracy across both classes with no adjustments, as suggested in \cite{liang2022metashift}.

\subsection{The Concept Sets used in CLIP-based CBMs and Human Annotation CBM}
\label{app:clip_cbms_concepts}
\textbf{Waterbirds.} \cite{koh2020concept} provide 112 visual attributes specifically related to the appearance of birds, not the backgrounds.
As described in \cref{app:CUB_anno}, these attributes have been annotated by humans.
We include these concept annotations ($\mathbf{c}$) corresponding to each image and generate data tuples $\{(\mathbf{x}^{(j)}, y^{(j)}, \mathbf{c}^{(j)})\}_{j=1}^n$ corresponding to data pairs $\{(\mathbf{x}, y)\}_{j=1}^n$ of Waterbirds.
In \cref{tab:waterbird_result}, \textit{Human Anno. CBM} refers to the annotation-based CBM based on these 112 attributes and annotations. 
\textit{LLaVA-based CBM (CUB)} in \cref{tab:waterbird_result} denotes the CBM constructed using the LLaVA-annotated concepts.
Post-hoc CBM \cite{yuksekgonul2022post} also uses these 112 attributes.
On the other hand, Label-free CBM~\cite{oikarinen2023label} suggest employing GPT-3 to gather essential visual features for classifying 200 bird species in the CUB dataset.
They collect a list of 370 visual attributes, focusing solely on bird appearance.
We create prompts based on these 370 attributes and use LLaVA to annotate them.
\textit{LLaVA-based CBM (Label-free)} denotes the resulting CBM.

\textbf{Metashifts and ImageNet-Opener.} 
Label-free CBM uses GPT-3 to collect concept sets as suggested in Sec 3.1 in \cite{oikarinen2023label}.
For ImageNet-Opener, the concepts used for Label-free CBM are \{\textit{a blade for cutting open cans, a blunt end, a can, a chair, a comfortable grip, a computer, a counter, a cupboard, a desk, a durable construction, a fridge, a kitchen, a lever for opening cans, a long, thin blade, a mailbox, a metal construction, a pen, a person, a pointed end, a printer, a sharp point at one end, a slender, sword-like shape, ``a small handle'', ``a small, handheld tool'', a table, envelopes, knife, opener}\}.
For Metashifts, Label-free CBM uses \{\textit{
a bowl,
a cat bed,
a cat food bowl,
a cat toy,
a collar,
a dog tag,
a leash,
a litter box,
a person,
a round face,
a scratching post,
a tail,
a toy,
a water bowl,
four legs,
fur,
green or yellow eyes,
large, oval eyes,
mammal,
pointed ears}\}.

On the other hand, Post-hoc CBM uses ConceptNet~\cite{speer2017conceptnet} to collect concepts. 
For ImageNet-Opener, the concept sets Post-hoc CBM used are \{\textit{sharp, knife, opener}\}.
For Methashifts, they are \{\textit{playing dead,
 been shaved,
 a,
 fleas,
 penis,
 claws,
 hungry,
 domestic animal,
 feline,
 black,
 big heart,
 four legs,
 hair,
 sharp claws,
 faithful companion,
 teeth,
 sharp teeth,
 two ears,
 brains,
 nose,
 woman,
 flag,
 gray,
 brown,
 loyal friend,
 gossip,
 the tail,
 eyes,
 mammal,
 pet,
 paws,
 alive,
 whisker,
 nice friend,
 fur,
 chap,
 mean,
 canine,
 good friend,
 fun,
 legs,
 thirsty,
 one mouth}\}.

\subsection{Using LLaVA directly to Make Predictions}
\label{app:llava_direct}
For Waterbirds, when we directly query LLaVA (13B) with \textit{“Is the bird in the image a waterbird or a landbird? Answer shortly.”}, the average accuracy is \textbf{74.2\%} and the worst group accuracy is \textbf{90.6\%}. LLaVA's response is biased towards classifying birds as waterbirds, resulting in an accuracy of only \textbf{4\%} for landbirds with water backgrounds.

\subsection{The Impact of Filtering Threshold on CBM performance.}
\label{all:threshold}
In Metashift, setting a threshold of 0.2 results in the detection of 16 correlated keywords and the removal of 3 spurious concepts among 27 concepts from Step 1. Raising the threshold to 0.25 reduces correlated keywords to 11, with no change in removed concepts.
Further raising the threshold to 0.33 retains all 27 concepts, resulting in a performance drop, with a worst-group accuracy of \textbf{75.2\%} and an average accuracy of \textbf{80.6\%}.

\subsection{Label-free Interpretability Results}
\label{app:label-free}

\begin{figure}
\setlength{\columnsep}{3pt}
  \centering
\includegraphics[width=0.65\textwidth]{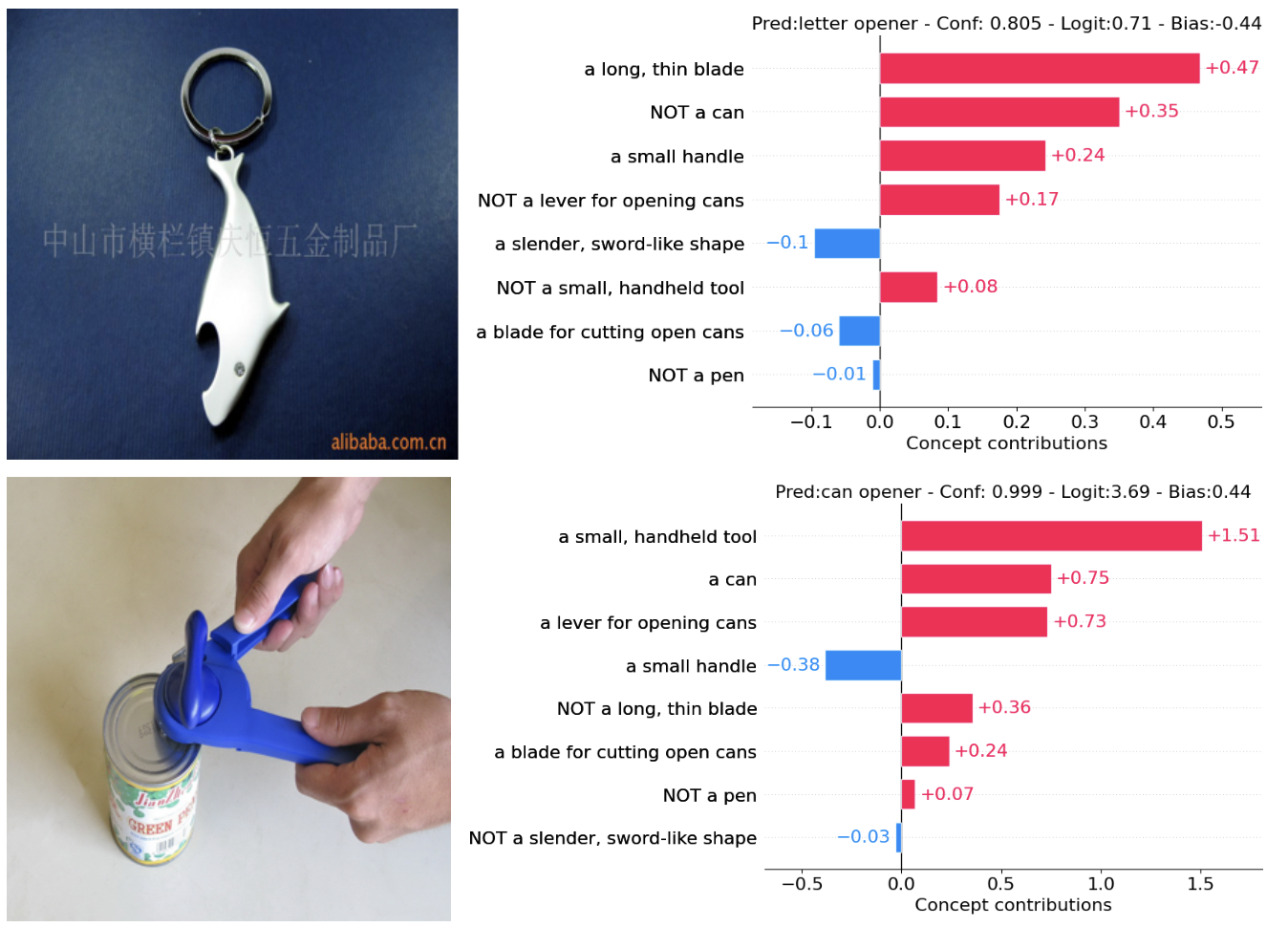}
   \caption{Label-free CBM's predictions rely on \textit{a can}.}
   \label{fig:label_free_ex}
\end{figure}

Figure~\ref{fig:label_free_ex} shows the inputs with explanations on the contribution of each concept to model's prediction. 
The image of \textit{can opener} lacking \textit{a can} tend to be misclassified (upper). Correct prediction occurs if \textit{a can} is present (lower).

\end{document}